\documentclass[letterpaper]{article} 
\usepackage{jmlr2e}
\usepackage{times}  
\usepackage{booktabs} 
\usepackage{url}
\usepackage{graphicx}
\usepackage{caption}
\usepackage{algorithm}
\usepackage[noend]{algpseudocode}
\usepackage{amsmath} 
\usepackage{amssymb}  
\usepackage{xcolor}
\usepackage{colortbl}
\usepackage{hyperref}
\usepackage{multicol}

\hypersetup{colorlinks = true,linkcolor=blue,citecolor = blue}
\usepackage{array,colortbl,ctable}
\usepackage{pythonhighlight}
\usepackage[toc,page]{appendix}
\newcommand{\name}{{MARGIN}}

\begin{document}

\title{\Large \bf MARGIN: Uncovering Deep Neural Networks using\\ Graph Signal Analysis}

\author{
	\name Rushil Anirudh\thanks{equal contribution} \email anirudh1@llnl.gov \\
	\addr Lawrence Livermore National Laboratory\\
	Livermore, CA,USA	
	\AND
	\name Jayaraman J. Thiagarajan \footnote[1] \\\email jjayaram@llnl.gov \\
	\addr Lawrence Livermore National Laboratory\\
	Livermore, CA,USA	
	\AND
	\name Rahul Sridhar \thanks{Work done as part of an internship at LLNL} \email rahul.sridhar@walmartlabs.com \\
	\addr Walmart Labs\\
	CA,USA
	\AND
	\name Peer-Timo Bremer \email bremer5@llnl.gov \\
	\addr Lawrence Livermore National Laboratory\\
	Livermore, CA,USA}
\maketitle

\begin{abstract}
 Interpretability has emerged as a crucial aspect of building trust in machine learning systems, aimed at providing insights into the working of complex neural networks that are otherwise opaque to a user. There are a plethora of existing solutions addressing various aspects of interpretability ranging from identifying prototypical samples in a dataset to explaining image predictions or explaining mis-classifications. While all of these diverse techniques address seemingly different aspects of interpretability, we hypothesize that a large family of interepretability tasks are variants of the same central problem which is identifying \emph{relative} change in a model's prediction. This paper introduces MARGIN, a simple yet general approach to address a large set of interpretability tasks MARGIN exploits ideas rooted in graph signal analysis to determine influential nodes in a graph, which are defined as those nodes that maximally describe a function defined on the graph. By carefully defining task-specific graphs and functions, we demonstrate that MARGIN outperforms existing approaches in a number of disparate interpretability challenges.

\end{abstract}

\section{Introduction}
With widespread adoption of deep learning solutions in science and engineering, obtaining post-hoc interpretations of the learned models has emerged as a crucial research direction. This is driven by a community-wide effort to develop a new set of meta-techniques able to provide insights into complex neural network systems, and explain their training or predictions. Despite being identified as a key research direction, there exists no well-accepted definition for interpretability. Instead, in different contexts, it may refer to a variety of tasks ranging from debugging models~\citep{ribeiro2016should}, to determining anomalies in the training data \citep{koh2017understanding}. While some recent efforts~\citep{lipton2016mythos, doshi2017roadmap} provide a more formal definition for interpretability as generating \textit{interpretable rules}, these focus on instance-level explanations, i.e.\ understanding how a network arrived at a particular decision for a single instance. In practice, interpretability covers a wider range of challenges, such as characterizing data distributions and separating hyper-planes of classifiers, identifying noisy labels during training, detecting adversarial attacks, or generating saliency maps for image classification. As discussed below, solutions to all such problems have been proposed each using custom tailored, task-specific approaches. For example, a variety of tools aim to explain which parts of an image are the most responsible for a prediction. However, these cannot be easily re-purposed to identify which samples in a dataset were most helpful or harmful to train a classifier.

\noindent
Instead, we argue that many existing interpretability techniques solve a variant of essentially the same task -- understanding \emph{relative} changes in the model's prediction, where the changes are either \emph{global} in nature, i.e., across an entire distribution or \emph{local}, i.e.,  within a single sample. In this paper, we propose the MARGIN (\textbf{M}odel \textbf{A}nalysis and \textbf{R}easoning using \textbf{G}raph-based \textbf{In}terpretability) framework, which directly applies to a wide variety of interpretability tasks. MARGIN poses each task as an {\it hypothesis} test and derives a measure of {\it influence} that indicates which parts of the data/model maximally support (or contradict) the hypothesis. More specifically, for each task we construct a graph whose nodes represent entities of interest, and define a function on this graph that encodes a hypothesis. For example, if the task is to determine which samples need to be reviewed in a dataset containing noisy labels, the domain is the set of samples, while the function can be local label agreement that measures how misaligned are the neighborhoods of the input samples (or their features) and their corresponding labels. Using graph signal processing~\citep{shuman2013emerging,sandryhaila2013discrete} one can then identify which nodes are essential to reconstructing the chosen function (hypothesis), which most likely will correspond to those with flipped labels. In order words, through a careful selection of graph construction strategies and hypothesis functions, this general procedure can be used to solve  a wide-range of post-hoc interepretability tasks.
%
%
\begin{figure*}[!h]
		\includegraphics[width=.99\linewidth,clip=True,trim=0 0 0 0]{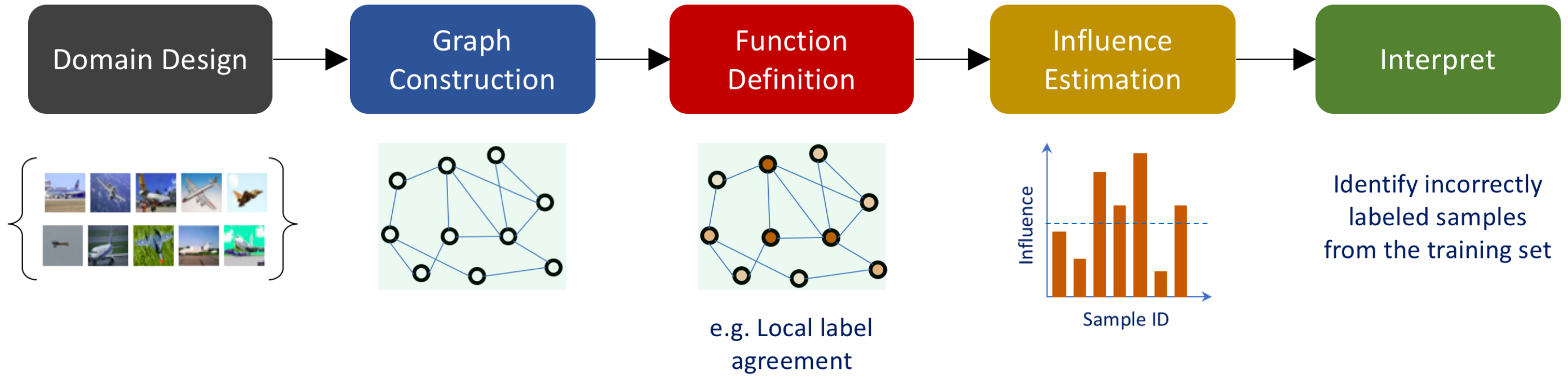}
	\caption{MARGIN - An overview of the proposed protocol for post-hoc interpretability tasks. In this illustration, we consider the problem of identifying incorrectly labeled samples from a given dataset. MARGIN identifies the most important samples that need to be corrected so that fixing them will lead to improved predictive models.}
\label{fig:margin}
\end{figure*}
%
This generic formulation, while extremely simple in its implementation, provides a powerful protocol to realize several meta-learning techniques, by allowing the user to incorporate rich semantic information, in a straightforward manner. In a nutshell, the proposed protocol is comprised of the following steps: (i) identifying the domain for interpretability (for e.g. intra-sample vs inter sample), (ii) constructing a neighborhood graph to model the domain, (for e.g. pixel space vs.\ latent space) (iii) defining an \textit{explanation function} at the nodes of the graph, (iv) performing graph signal analysis to estimate the \textit{influence} structure in the domain, and (v) creating interpretations based on the estimated influence structure. Figure \ref{fig:margin} illustrates the steps involved in MARGIN for \textit{a posteriori} interpretability.


\noindent \textbf{Overview:} Using different choices for graph construction and the explanation function design, we present five case studies to demonstrate the broad applicability of MARGIN for \textit{a posteriori} interpretability. First, in section \ref{sec:critics} we study a unsupervised problem of identifying samples which well characterize the underlying data distribution, referred to as prototypes and criticisms respectively \citep{kim2016examples}. We show that the MARGIN is highly effective at characterizing data distributions and can shed light into the regimes where classifier performance can suffer. In section \ref{sec:saliency}, we obtain pixel-level explanations from an image classifier using MARGIN, without the need to access the model internals, i.e., black-box and show that the inferred feature importance estimates are meaningful. In section \ref{sec:noisy}, we employ MARGIN to identify label corruptions in the training data and demonstrate significant improvements over popular approaches such as influence functions. In section \ref{sec:decision}, we illustrate the application of MARGIN in analyzing pre-trained classifiers and identifying the most influential samples in describing the decision surfaces, akin to memorable examples in continual learning~\citep{pan2020continual}. Finally, in section \ref{sec:adversarial} we extend two recently proposed statistical  techniques to detect adversarial examples from harmless examples, and demonstrate that incorporating them inside MARGIN improves their discriminative power significantly.


\section{Related Work}
We outline recent works that are closely related to the central framework, and themes around MARGIN. Papers pertinent to individual case studies are identified in their respective sections.

\noindent Our goal in this paper is to design a core framework that is capable of being re-purposed to interpretability tasks, ranging from explaining decisions of a predictive model, detecting outliers to identifying label corruptions in the training data. While post-hoc explanation methods are the \textit{modus-operandi} in interpreting the decisions of a black box model, their scope has widened significantly in the recent years. For example, popular sensitivity analysis such as LIME~\citep{ribeiro2016} and SHAP~\citep{lundberg2017unified} or gradient-based methods such as Saliency Maps~\citep{simonyan2013deep}, Integrated Gradients~\citep{sundararajan2017axiomatic}, Grad-CAM~\citep{selvaraju2017grad}, DeepLIFT~\citep{shrikumar17a} and DeepSHAP~\citep{lundberg2017unified} are routinely used to produce sample-wise, local explanations by measuring the sensitivity of the black-box to perturbations in the input features~\citep{fong2017interpretable}. Despite their wide-spread use, they cannot be readily utilized to obtain dataset-level explanations, e.g., which are the most influential examples in a dataset for a given test sample, or to detect distribution shifts~\citep{thiagarajan2020accurate}. On the other hand, in \citep{koh2017understanding}, the authors proposed a strategy to select influential samples by extending ideas from robust statistics, which was shown to be applicable to a variety of scenarios. However, such methods cannot be used for obtaining feature importance estimates. Another important challenge with most existing post-hoc explanation techniques is their computational complexity. In contrast, MARGIN leverages the generality of graph structures to generate explanations in a scalable fashion, and through of use of appropriate hypothesis functions can support a large-class of interpretations. 

In a nutshell, MARGIN reposes the problem of generating explanations as an influential node selection problem, wherein the node can correspond to a sample-level or feature-level explanations and the influence is measured based on a hypothesis function. Defining suitable objectives for detecting influential features in an image or influential samples in a dataset has been an important topic of research in explainable AI. For example, CXPlain~\citep{schwab2019cxplain} and Attentive Mixture of Experts~\citep{schwab2019granger} utilize a Granger-causality based objective to quantify feature importances. In addition, prediction uncertainties~\citep{chakraborty2017interpretability} or even loss estimates~\citep{thiagarajan2020accurate} have been widely adopted to characterize vulnerabilities of a trained model. Note that, MARGIN can directly use any of these objectives to choose the most relevant explanations. In this paper, we consider a variety of interpretability tasks and recommend suitable hypothesis functions for each of the tasks. 

Since MARGIN relies on ideas from graph signal processing (GSP) to select the most relevant explanations, we briefly review existing work in this area. Broadly, there are two classes of approaches in GSP -- one that builds on spectral graph theory using the graph Laplacian matrix \citep{shuman2013emerging}, and the other based on algebraic signal processing that builds upon the graph shift operator \citep{sandryhaila2013discrete}. While both are applicable to our framework, we adopt the latter formulation. Our approach relies on defining a measure of influence at each node, which is related to sampling of graph signals. This is an active research area, with several works generalizing ideas of sampling and interpolation to the domain of graphs, such as \citep{chen2015discrete,pesenson2008sampling,gadde2014active}. 




\section{A Generic Protocol for Interpretability}
\label{sec:framework}
In this section, we provide an overview of the different steps of MARGIN and describe the proposed influence estimation technique in the next section.

\noindent \textbf{Domain Design and Graph Construction:} The domain definition step is crucial for the generalization of MARGIN across different scenarios. In order to enable instance-level interpretations (e.g. creating saliency maps), a single instance of data, possibly along with its perturbed variants, will form the domain; whereas a more holistic understanding of the model can be obtained (e.g. extracting prototypes/criticisms) by defining the entire dataset as the domain. Regardless of the choice of domain, we propose to model it using nearest neighbor graphs, as it enables a concise representation of the relationships between the domain elements.

More specifically, given the set of samples $\{\mathbf{x}_i\}$, we construct a $k$-nearest neighbor domain graph that captures local geometry of the data samples. The metric for graph construction (that determines neighborhoods/edges) can arise from prior knowledge about the domain or designed based on latent representations from pre-trained models. For example, if we use the latent features from AlexNet \citep{alexnet}, the resulting graph respects the distance metric inferred by AlexNet for image classification. Though the difficulty in choosing an appropriate $k$ for designing robust graphs is well known, designing better graphs is beyond the scope of this paper. In our experiments, we find that our results are not very sensitive to the choice of $k$.

Formally, an undirected weighted graph is represented by the triplet $\mathcal{G} = (\mathcal{V}, \mathcal{E}, \mathbf{W})$, where $\mathcal{V}$ denotes the set of nodes, $\mathcal{E}$ denotes the set of edges and $\mathbf{W}$ is an adjacency matrix that specifies the weights on the edges, where $\textbf{W}_{n,m}$ corresponds to the edge weight between nodes $v_n$ and $v_m$. Let $\mathcal{N}_{n} = \{m | \mathbf{W}_{n,m} \neq 0\} $ define the neighborhood of node $v_n$, i.e. the set of nodes connected to it. The normalized graph Laplacian, $\mathbf{L}$, is then constructed as $\mathbf{I} - \mathbf{D}^{-1/2}\mathbf{W} \mathbf{D}^{-1/2}$, where $\mathbf{D}_{nn} = \sum_m \mathbf{W}_{n,m}$ is the degree matrix and $\mathbf{I}$ denotes the identity matrix.

\noindent \textbf{Explanation Function Definition:} A key component of MARGIN is to construct an explanation function that measures how well each node in the graph supports the presented hypothesis. The function acts on the $n^{th}$ vertex of the graph as $f(n)$, which is a real-valued function that takes the properties of vertex $v_n$ as input, for all $N$ vertices in the graph $\mathcal{G}$. This function is also referred to as the graph signal defined on the graph domain. We expect this function to capture properties of the explanation that are deemed important. Let us illustrate this process with an example -- in order to create saliency maps for image classification, one can build a graph where each node corresponds to a potential explanation (i.e. a subset of pixels), while the edges can measure how likely can two explanations produce similar predictions. In such a scenario, one can hypothesize that an \emph{ideal} explanation will be \textit{sparse}, in terms of the number of pixels, since that is more interpretable. Consequently, the size of an explanation can be used as the function. Section \ref{sec:results} will present a more detailed discussion.

\noindent \textbf{Influence Estimation:} This is the central analysis step in MARGIN for obtaining influence estimates at the nodes of $\mathcal{G}$, that can reveal which nodes can maximally describe the variations in the chosen explanation function. Implicitly, this step can be viewed as a \textit{soft-}sample selection strategy with respect to the structure induced by the domain graph. We propose to perform this estimation using tools from graph signal analysis. Section \ref{sec:influence} describes the proposed algorithm for influence estimation.

\noindent \textbf{From Influence to Interpretation:} Depending on the hypothesis chosen for \textit{a posteriori} analysis, this step requires the design of an appropriate strategy for transferring the estimated influences into an interpretable explanation.

\section{Proposed Influence Estimation}
\label{sec:influence}
Given a nearest neighbor graph $\mathcal{G}$ along with an explanation function $\mathbf{f}$, we propose to employ graph signal analysis to estimate node influence scores. Before we describe the algorithm, we will present a brief overview of the preliminaries.

\noindent \textbf{Definitions:} We use the notation and terminology from \citep{sandryhaila2013discrete} in defining an operator analogous to the \emph{time-shift} or \emph{delay} operator in classical signal processing. The function $\mathbf{f}$ assigns a scalar value $f$, to each vertex as defined earlier, as a result the entire function is written as $\mathbf{f} = \{f_1,f_2,\dots,f_N\}$, is a collection of scalar values at each vertex, ordered according to the same order of vertices in the graph. When the graph does not have any special structure (i.e., it is Euclidean), $\mathbf{f}$ is a vector valued function. We consider the simplest scenario here where the function only takes a scalar value at each node, however more general cases maybe considered where the value at each node is multi-dimensional, in which case $\mathbf{f}$ is a matrix.  During a graph shift operation, the function $\mathbf{f}(n)$ at node $v_n$ is replaced by a weighted linear combination of its neighbors: $\hat{\mathbf{f}} = \mathbf{A}\mathbf{f}$, where $\mathbf{A}$ is the graph shift operator, which is the simplest, non-trivial graph filter. Commonly used choices for $\mathbf{A}$ include the adjacency matrix $\mathbf{W}$, transition matrix $\mathbf{D}^{-1}\mathbf{W}$ and the graph Laplacian $\mathbf{L}$.

The set of eigenvectors of the graph shift operator is referred to as the graph Fourier basis, $\mathbf{A} = \mathbf{U} \mathbf{\Lambda} \mathbf{U}^T$, where $\mathbf{U} \in \mathbb{R}^{N\times N}$, and the Fourier transform of a signal $\mathbf{f} \in \mathbb{R}^N$ is defined as $\mathbf{U}^T \mathbf{f}$. The ordered eigenvalues corresponding to these eigenvectors represent frequencies of the signal, with $\lambda_1$ to $\lambda_N$ representing the smallest to largest frequencies. The notion of frequency on the graph corresponds to the rate of change of the function across nodes in a neighborhood. A higher change corresponds to a high frequency, while a smooth variation corresponds to a low frequency. In this context, the graph filtering using a graph shift operator corresponds to a \emph{low-pass} filter that dispenses high frequency components in the function. Similarly, a simple \emph{high-pass} filter can be easily designed as $\hat{\mathbf{f}_{h}} = \mathbf{f} - \hat{\mathbf{f}}$.

\begin{algorithm}[!htb]
	\caption{MARGIN's simple influence estimation}
	\label{alg:select}
	\begin{algorithmic}[1]
		\Procedure{MARGIN-Influence}{$\mathbf{X}$,$G$, $\mathbf{f}$}\Comment{Domain, Graph and explanation function}
			\State Construct graph shift operator $\mathbf{A}$ from $\mathbf{X}$
			\State $\hat{\mathbf{f}} = \mathbf{f} - \mathbf{A}\mathbf{f}$ \Comment{High pass spectral filter}
			\For{$i \in \mathcal{V}$}\Comment{Iterate over all the nodes of the graph $\mathcal{G}$}
				\State Compute $I(i)  = ||\hat{\mathbf{f}}(i)||^2_2~~ \forall i \in \mathcal{V}$
			\EndFor
			\State \textbf{return} $I(i) \forall i \in \mathcal{V}$ \Comment{Influence score for each node}
		\EndProcedure
	\end{algorithmic}
\end{algorithm}

\noindent \textbf{Algorithm:} The overall procedure to obtain influence scores at the nodes of $\mathcal{G}$ can be found in Algorithm \ref{alg:select}. Intuitively, we design a high-pass filter that eliminates the low frequency content and retains the signal energy only at those nodes that characterize the extreme variations of the function. Following the high-pass filtering step, the influence score at a node is estimated as the magnitude of the filtered function value at that node:
\begin{equation}
\label{eq:graph_influence}
I(i) = ||\hat{\mathbf{f}_{h}}(i)||^2_2~~ \forall i \in \mathcal{V},
\end{equation}where $\hat{\mathbf{f}_{h}}$ corresponds to the high-pass filtered version of $\mathbf{f}$. Interestingly, we find that analyzing the high frequency components of the explanation function often leads to a sparse influence structure, indicating the presence of multiple local optima that corroborate the hypothesis. Conversely, the influence structure obtained from low frequency components is typically dense and hence requires additional processing to qualify regions of disagreement.

\noindent \textbf{Sensitivity to graph construction} A critical step in MARGIN is the graph construction process for datasets that do not naturally have a graph structure. In this work, we rely on a simple nearest neighbor graph for construction which can vary depending on the size of the neighborhood. This is a hyper parameter that must be set with validating examples, and in all our case studies we found a neighborhood size of 20-40 to be quite good in terms of computational efficiency in constructing the graph -- the exact value can vary depending on the nature of the application.

This directly influences the quality of low pass filtering of a graph signal similar to the case in Euclidean signal processing in choosing a size of the window. As the neighborhood size increases, the filtering at each node becomes more aggressive since it averages the across several neighboring nodes, while for a small neighborhood the smoothing may not have any effect at all. MARGIN is agnostic to the type of graph construction used, since it ultimately only relies on the graph filtering process. As a result it is applicable to other graph constructions such as Reeb graphs \citep{pascucci2007robust} or $\beta-$skeletons. However, the choice of metric in the graph construction is critical, since it can determine how the neighborhood is determined which ultimately determines influence of each node.

\section{Case Studies}
Considering MARGIN is very generic in nature, it is easy applicable to a wide variety of interpretability tasks. In this section we illustrate this felxibility on several example tasks. Table \ref{table:use-case-setup} shows the domain design, graph construction, and function definition choices made for different use cases. Note in each case study, we construct a $k$-nearest-neighbor graph followed by the application of MARGIN with the main difference is in how the nodes of the graph are defined, followed by the type of function that is defined at each node. 

\begin{table*}[!htb]
	\centering
	\caption{Using MARGIN to solve different commonly encountered interpretability tasks.}
	\label{tab:explain}
	\begin{tabular}{|p{1.0in}|p{1.3in}|p{1.3in}|p{1.0in}|p{1.4in}|}
		\hline
		\small
		\cellcolor{gray!30}\textbf{Task} & \cellcolor{gray!30}\textbf{Domain} & \cellcolor{gray!30}\textbf{Nodes in $\mathcal{G}$} & \cellcolor{gray!30}\textbf{Function} & \cellcolor{gray!30}\textbf{Explanation Modality} \\
		\hline
		Prototypes/ Criticisms & Complete dataset 			& Samples  					& MMD (Global,Local)  	& Sample sub-selection  \\
		 &  			&   					&  	&   \\
		Explain prediction 		& Single image 					& Explanations  		& Sparsity  						& Saliency maps  \\
		&  			&   					&  	&   \\
		Detect noisy-labels 	& Complete dataset 			& Samples  					& Local label-agreement & Samples to fix \\
		&  			&   					&  	&   \\
		Detect adversarial-attacks 	& Attacks/Noisy samples & Perturbed samples & MMD(Global)  				& Attack statistics \\
		&  			&   					&  	&   \\
		Study discrimination  & Complete dataset 			& Samples 					& Local label-agreement & Confusing samples \\
		\hline
	\end{tabular}
\label{table:use-case-setup}
\end{table*}

\label{sec:results}
\subsection{Case Study I - Prototypes and Criticisms}
\label{sec:critics}
A commonly encountered problem in interpretability is to identify samples that are prototypical of a dataset, and those that are statistically different from the prototypes (called criticisms). Together, they can provide a holistic understanding about the underlying data distribution. Even in cases where we do not have access to the label information, we seek a hypothesis that can pick samples which are representatives of their local neighborhood, while emphasizing statistically anomalous samples. One such function was recently utilized in \citep{kim2016examples} to define prototypes and criticisms, and it was based on Maximum Mean Discrepancy (MMD).

\noindent \textbf{Formulation:} Following the general protocol in Figure \ref{fig:margin}, the domain is defined as the complete dataset, along with labels if available. Since this analysis does not rely on pre-trained models, we construct the neighborhood graph based on the Euclidean distance using $k=25$ nearest neighbors. Inspired by \citep{kim2016examples}, we define the following explanation function: For each sample $\mathbf{x}_i$, we remove the chosen sample and all its connected neighbors from the graph to construct the set $\bar{\mathcal{X}}_i = \{\mathbf{x}_j, j \notin ( i \cup \mathcal{N}_i)\}$, and estimate the function at the $i^{th}$ node as $f(i) = \mbox{MMD}(\bar{\mathcal{X}}_i$,$\bar{\mathcal{X}}_i \cup \mathbf{x}_i$). MMD gives us a way to measure the difference between two distributions, and since we artificially construct the two distributions by removing a single sample, we are able to determine the importance of an individual sample (and its neighbors) within the dataset using MARGIN. Let $k:\mathcal{X}\times \mathcal{X}\mapsto \mathrm{R}$ be a kernel such as the radial basis function (RBF) kernel, and $\mathcal{X} = \bar{\mathcal{X}}_i \cup \mathbf{x}_i$, then we can use the approximation for MMD given in (c.f. eqn (5) in \cite{kim2016examples}) as : 
\begin{equation}
\label{eq:mmd-critic}
\mbox{MMD}(\mathbf{x}_i) = \frac{1}{|\bar{\mathcal{X}}_i|}\sum_{\mathbf{x}_m \in \bar{\mathcal{X}}_i} k(\mathbf{x}_i,\mathbf{x}_m) + \frac{1}{|\mathcal{X}|}\sum_{\mathbf{x}_j \in \mathcal{X}} k(\mathbf{x}_i,\mathbf{x}_j)
\end{equation}

In cases of labeled datasets, the kernel density estimates for the MMD computation are obtained using only samples belonging to the same class. We refer to these two cases as \textit{global} (unlabeled case) and \textit{local} (labeled case) respectively. The hypothesis is that the regions of criticisms will tend to produce highly varying MMD scores, thereby producing high frequency content, and hence will be associated with high MARGIN scores. Conversely, we find that the samples with low MARGIN scores correspond to prototypes since they lie in regions of strong agreement of MMD scores. More specifically, we consider all samples with low MARGIN scores (within a threshold) as prototypes, and rank them by their actual function values. In contrast to the greedy inference approach in \citep{kim2016examples} that estimates prototypes and criticisms separately, they are inferred jointly in our case.

\noindent \textbf{Experiment Setup and Results:} We evaluate the effectiveness of the chosen samples through predictive modeling experiments with the idea that the most helpful samples should result in a good classifier, whereas a the most unhelpful/confusing samples should result in a poor classifier. In other words, we evaluate how well a classifier can generalize when trained only using criticisms or prototypes. We use the USPS handwritten digits data for this experiment, which consists of 9,298 images belonging to 10 classes. We use a standard train/test split for this dataset, with 7,291 training samples and the rest for testing. For fair comparisons with \citep{kim2016examples}, we use a simple 1-nearest neighbor classifier. As described earlier, we consider both unsupervised (\textit{global}) and supervised (\textit{local}) variants of our explanation function for sample selection.

\begin{figure*}[!htb]

	\includegraphics[width=.99\linewidth]{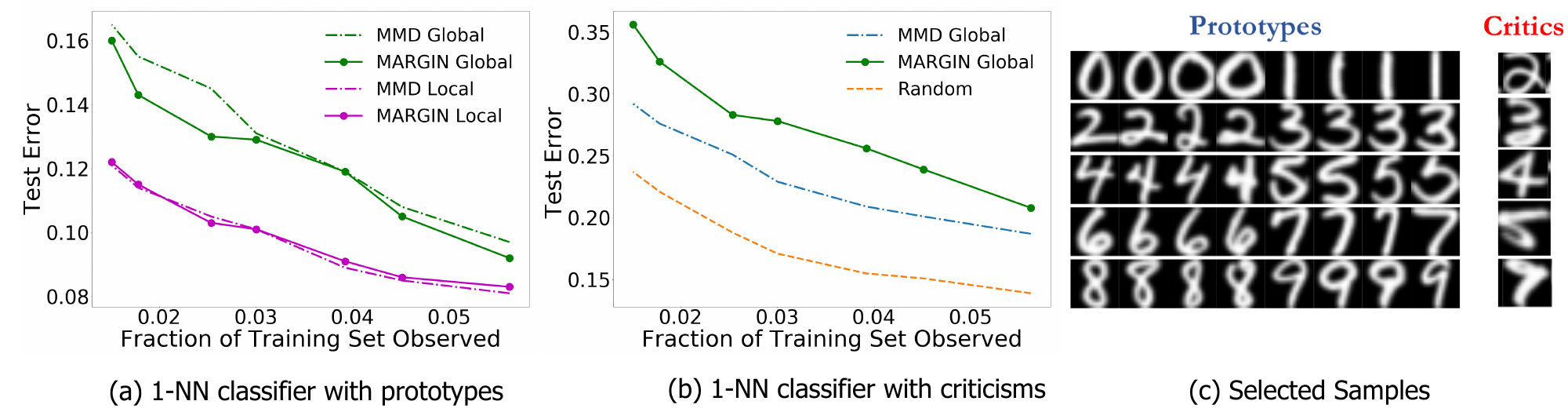}
	\caption{Using MARGIN to sample prototypes and criticisms. In this experiment, we study the generalization behavior of models trained solely using prototypes or criticisms.}
	\label{fig:noise}
\end{figure*}

We expect the prototypical samples to be the most helpful in predictive modeling, i.e., good generalization. In Figure \ref{fig:noise}(a), we observe that the prototypes from MARGIN perform competitively in comparison to the baseline technique. More importantly, MARGIN is particularly superior in the global case, with no access to label information. On the other hand, criticisms are expected to be the least helpful for generalization, since they often comprise boundary cases, outliers and under-sampled regions in space. Hence, we evaluate the test error using the criticisms as training data. Interestingly, as shown in Figure \ref{fig:noise}(b), the criticisms from MARGIN achieve significantly higher test errors in comparison to samples identified using \textit{MMD-critic} based optimization in \citep{kim2016examples}. Furthermore, examples of the selected prototypes and criticisms from MARGIN are included in Figure \ref{fig:noise}(c).

\subsection{Case Study II - Explanations for Image Classification}
\label{sec:saliency}
Generating explanations for predictions is crucial to debugging black-box models and eventually building trust. Given a model, such as a deep neural network, that is designed to classify an image into one of $r$ classes, a plausible \textit{explanation} for a test prediction is to quantify the importance of different image regions to the overall prediction, i.e. produce a saliency map. We posit that perturbing the salient regions should result in maximal changes to the prediction. In addition, we expect \textit{sparse} explanations to be more interpretable. In this section, we describe how MARGIN can be applied to achieve both these objectives.

\noindent \textbf{Formulation:} Since we are interested in producing explanations for instance-level predictions using MARGIN, the domain corresponds to a possible set of explanations for an image. Note that, the space of explanations can be combinatorially large, and hence we adopt the following greedy approach to construct the domain. We run the SLIC algorithm \citep{slic} with varying number of superpixels, say $\{50, 100, 150, 200, 250, 300\}$, and define the domain as the union of superpixels from all the independent runs. In our setup, each of these superpixels is a plausible explanation and they become the nodes of $\mathcal{G}$. The edge between nodes $m$ and $n$ of this graph is defined based on the relative importance of each super-pixel, i.e., $e_{mn} = |~|p_j(\mathcal{I})-p_j(\mathcal{I}_m)| - |p_j(\mathcal{I})-p_j(\mathcal{I}_n)|~|$, where $\mathcal{I}$ is the original image, and $\mathcal{I}_m$ is the image with the $m^{th}$ super-pixel masked out, and $p_j(~)$ extracts the softmax scores for the $j^{th}$ class in the image. This relative importance captures how two super-pixels are related in terms of the predictive model, which is related to a causal objective that is used in CXPlain \citep{schwab2019cxplain}.

\begin{figure*}[!htb]
	\centering
	\includegraphics[width=.95\linewidth,clip=True,trim=0 0 0 0]{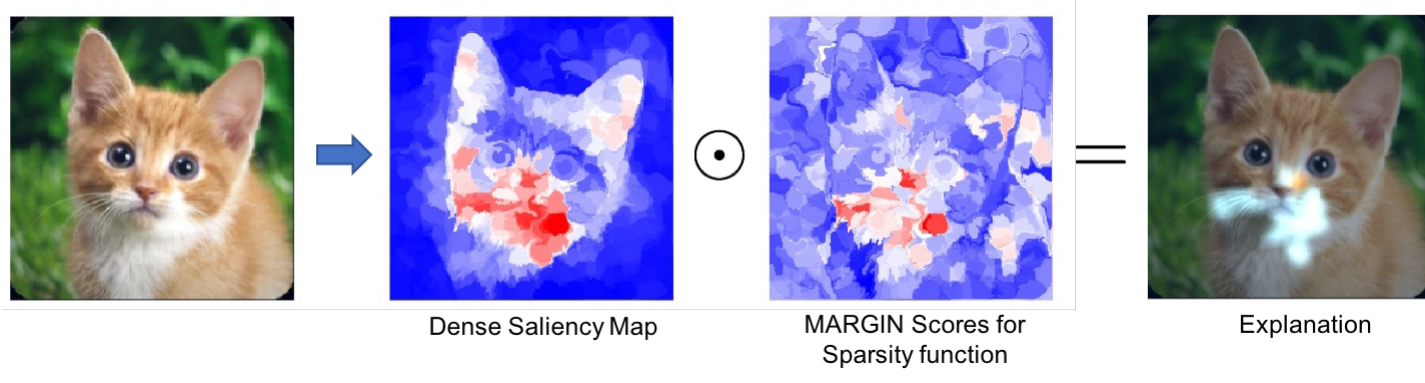}
	\vspace{-10pt}
	\caption{We show the entire process of constructing the saliency map for one particular image (Tabby Cat) from ImageNet. From left to right: original image, (dense) saliency map $S$, sparsity map $I$, and finally the explanation from MARGIN, $S_{final}$.}
	\label{fig:saliency_tabby}
	\vspace{-5pt}
\end{figure*}

For each of the explanations (super-pixels) $m$, we mask its pixels in the image and use the pre-trained model to obtain a measure of its saliency as before as $|p_j(\mathcal{I}) - p_j(\mathcal{I}_m)|$. Using these estimates, we obtain pixel-level saliency, $S$, as a weighted combination of their saliency from different superpixels (inversely weighted by the superpixel size). This dense saliency is similar to previous approaches such as \citep{zeiler2014visualizing, zhou2014object}.

Note that, this saliency estimation process did not impose the sparsity requirement. Hence, we use MARGIN to obtain influence scores based on their sparsity. The explanation function at each node is defined as the ratio of the size of the superpixel corresponding to that node and the size of the largest superpixel in the graph. Intuitively, MARGIN finds the sparsest explanation for different level sets of the saliency function. Subsequently, we compute pixel-level influence scores, $I$, as a weighted combination of their influences from different superpixels. The overall saliency map is obtained as $S_{final} = S \odot I$, where $\odot$ refers to the Hadamard product.

\begin{figure*}[!htb]
	\centering
	\includegraphics[width=.95\linewidth,clip=True,trim=0 0 0 0]{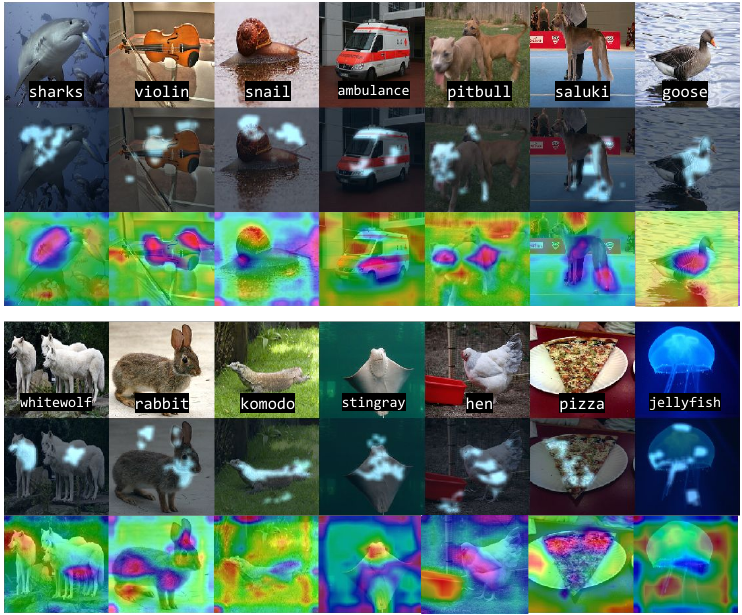}
	\vspace{-10pt}
	\caption{Our approach identifies the most salient regions in different classes for image classification using AlexNet. From top to bottom: original image, MARGIN's explanation overlaid on the image, and Grad-CAM's \citep{gradcam} explanation. Note our approach yields highly specific, and sparse explanations from different regions in the image for a given class.}
	\label{fig:saliency_maps_add}
\end{figure*}

\noindent \textbf{Experiment Setup and Results:} Using images from the ImageNet database \citep{russakovsky2015imagenet}, and the AlexNet \citep{alexnet} model, we demonstrate that MARGIN can effectively produce explanations for the classification. Figure \ref{fig:saliency_tabby} illustrates the process of obtaining the final saliency map for an image from the \textit{Tabby Cat} class. Interestingly, we see that the mouth and whiskers are highlighted as the most salient regions for its prediction. Figure \ref{fig:saliency_maps_add} shows the saliency maps from MARGIN for several other cases. For comparison, we show results from Grad-CAM \citep{gradcam}, which is a white-box approach that accesses the gradients in the network. We find that, using only a black-box approach, MARGIN produces explanations that strongly corroborate with Grad-CAM and in some cases produces more interpretable explanations. For example, in the case of an \textit{Ice Cream} image, MARGIN identifies the ice cream, and the spoon, as salient regions, while Grad-CAM highlights only the ice cream and quite a few background regions as salient. Similarly, in the case of a \textit{fountain} image, MARGIN highlights the fountain, and the sky, while Grad-CAM highlights the background (trees) slightly more than the fountain itself, which is not readily interpretable.
%

\subsection{Case Study III - Detecting Incorrectly Labeled Samples}
\label{sec:noisy}
An increasingly important problem in real-world applications is concerned with the quality of labels in supervisory tasks. Since the presence of noisy labels can impact model learning, recent approaches attempt to compensate by perturbing the labels of samples that are determined to be high-risk of being corrupted, or when possible have annotators check the labels of those high-risk samples. In this section, we propose to employ MARGIN to recover incorrectly labeled samples. In particular, we consider a binary classification task, where we assume $\beta$\% of the labels are randomly flipped in each class. In order to identify samples which were incorrectly labeled, we select samples with the highest MARGIN score, followed by simulating a human user correcting the labels for the top $K$ samples. Ideally, we would like $K$, the number of samples checked by the user, to be as small as possible.

\noindent \textbf{Formulation:} Similar to Case Study I, the entire dataset is used to define the domain.  Since we expect the flips to be random, we hypothesize that they will occur in regions where the labels of corrupted samples are different from their neighbors. Instead of directly using the label at each node as the explanation function, we believe a more smoothly varying function will allow us to extract regions of high frequency changes more robustly. As a result, we propose to measure the level of \emph{distrust} at a given node, by measuring how many of its neighbors disagree with its label:
\begin{equation}
\label{eq:distrust}
\mathbf{f}(i) = 1 - \frac{\sum_{j\in \mathcal{N}_i}L(j,i)}{|\mathcal{N}_i|},
\end{equation}where $L(j,i)$ is $1$ only if nodes $j$ and $i$ share the same label; $|.|$ denotes the cardinality of a set.

\begin{figure*}[t]
	\centering
		\includegraphics[width=.95\linewidth,clip=True,trim=0 0 0 0]{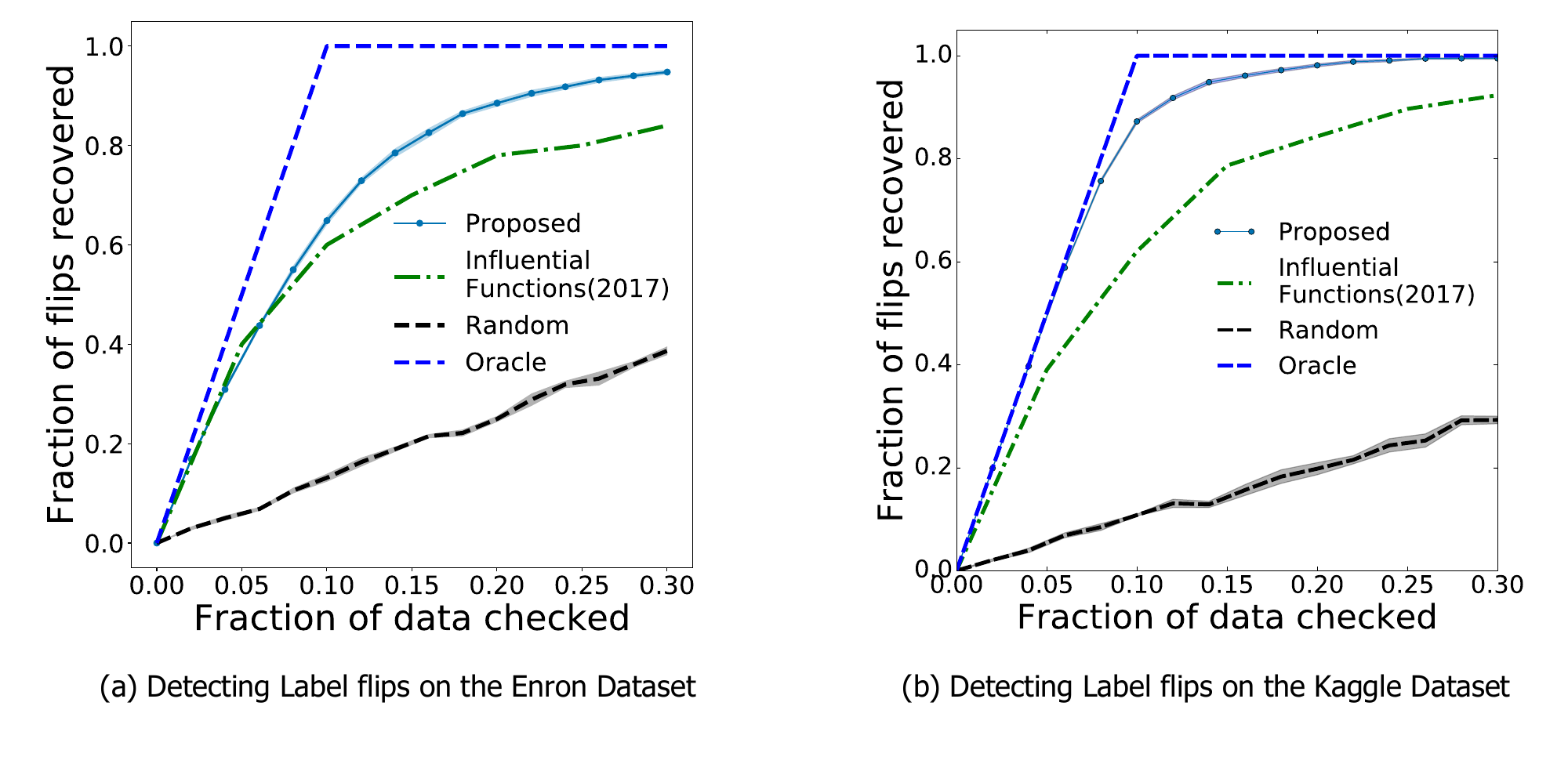}
	\caption{MARGIN can be used to find samples with incorrect labels efficiently, much better than competing influence sampling based approaches. The 'Oracle' here is the best case scenario, where the samples checked are exactly the ones that are corrupted.}
	\label{fig:corruption}
	\vspace{-10pt}
\end{figure*}

\noindent \textbf{Experiment Setup and Results:} We perform our experiments on two datasets: (a) the Enron Spam Classification dataset \citep{metsis2006spam}, containing $4138$ training examples, with an imbalanced class split of around 70:30 (non-spam:spam), and (b) 3000 random images from Kaggle dog v cat classification dataset with almost equal number of images from each class\footnote{\url{https://www.kaggle.com/c/dogs-vs-cats/data}}. Following standard practice, we randomly corrupt the labels of $10\%$ of the samples. For the Enron Spam dataset, we extracted bag-of-words features of $500$ dimensions corresponding to the most frequently occurring words. We observed these features to be noisy, so we use a simple PCA pre-processing step to reduce the dimensionality of the data down to 100. For Kaggle, we use penultimate features from AlexNet \cite{alexnet} in order to construct a neighborhood graph. In both cases we use $k=20$ as the number of neighbors for this purpose, we observed stable performance even when $k=30$ or $k=40$. The use of features instead of the data directly has become standard practice in several applications as it reduces the dimensionality of the data, while also providing a more semantically meaningful notion of neighborhood. We report average results from 10 repetitions of the experiment. 

\noindent We compare our approach with three baselines: \noindent {\it(i) Influence Functions:} We obtain the most influential samples using Influence Functions \citep{koh2017understanding}. {\it(ii) Random Sampling} {\it(iii) Oracle:} The best case scenario, where the number of labels corrected is equal to the number of samples observed. Following \citep{koh2017understanding}, we vary the percentage of influential samples chosen, and compute the \textit{recall} measure, which corresponds to the fraction of label flips recovered in the chosen subset of samples.

As seen in Figure \ref{fig:corruption}, we see that our method is nearly $10$ percentage points better than the state-of-the-art Influence Functions, achieving a recall of nearly $0.95$ by observing just 30\% of the samples. This difference is further improved when observing a balanced dataset like the Kaggle dogs v cats, as seen in Figure \ref{fig:corruption}(b) where MARGIN outperforms Influence functions significantly. On examining how MARGIN picks the samples, we see a clear trend which indicates a strong preference for samples that lie farther away from the classification boundary. In other words, this corresponds strongly to correcting the least number of samples which can lead to the most gain in validation performance when using a trained model.

\begin{figure*}[t]
	\centering
	\includegraphics[width=.95\linewidth,clip=True,trim=0 0 0 0]{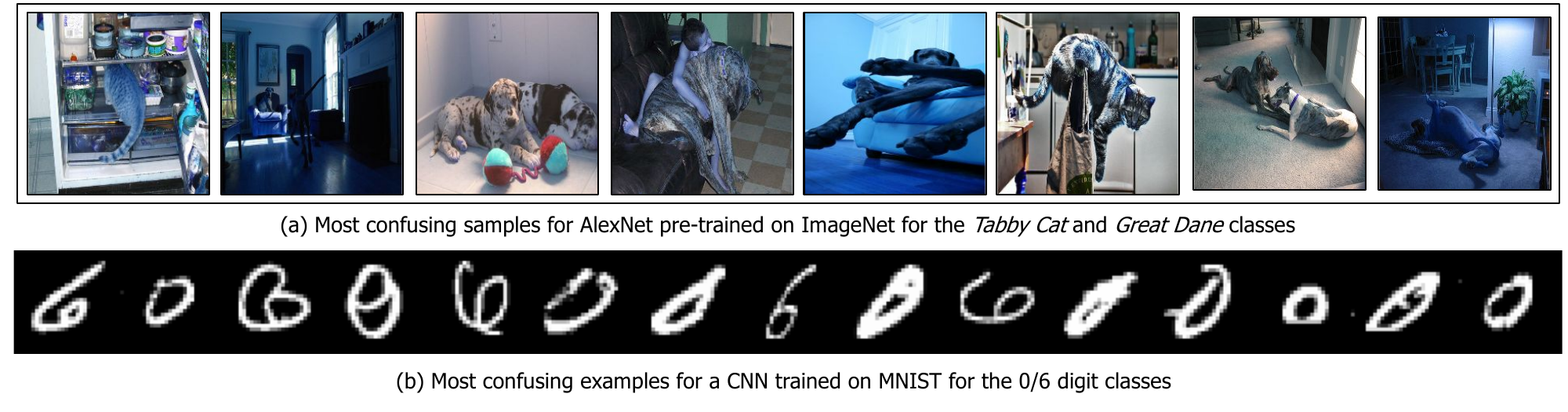}
	\caption{Using MARGIN to sample near decision boundaries.}
	\label{fig:decsur}
\end{figure*}

 \subsection{Case Study IV - Interpreting Decision Boundaries}
\label{sec:decision}
While studying black-box models, it is crucial to obtain a holistic understanding of their strengths, and more importantly, their weaknesses. Conventionally, this has been carried out by characterizing the decision surfaces of the resulting classifiers.  In this experiment, we demonstrate how MARGIN can be utilized to identify samples that are the most confusing to a model, or more precisely those examples which are likely to be mis-classified by a pre-trained classifier. By definition these are images that are closest to the decision boundary inferred by the classifier.

\noindent \textbf{Formulation:} In order to adopt MARGIN for analyzing a specific model, we construct a nearest neighbor graph ($k=30$) using latent representations inferred from the pre-trained classifier in consideration. This achieves two things -- it gives us a semantic similarity measure as interpreted by the classifier, i.e., which similarities are considered important for the classification task. More importantly for this case study, this automatically distills the information regarding confusing samples into the graph that is constructed, since these samples are likely to be in regions of the neighborhood with high prediction uncertainty. Next, since the decision surface characterization is similar to Case Study III, we use the local label agreement measure in (\ref{eq:distrust}) as the explanation function. This disagreement between the function and the neighborhood shows up as high frequency information which is exploited by MARGIN to identify the decision surface.

\noindent \textbf{Experiment Setup:} We perform an experiment on 2-class datasets extracted from ImageNet and MNIST. More specifically, in the case of ImageNet, we perform decision surface characterization on the classes \textit{Tabby Cat} and \textit{Great Dane}. We used the features from a pre-trained AlexNet's\citep{alexnet} penultimate layer to construct the graph. For the MNIST dataset, we considered data samples from digits `0' and `6', and we used the latent space produced using a convolutional neural network for the analysis. A selected subset of samples characterizing the decision surfaces of both datasets are shown in Figure \ref{fig:decsur}.

 \noindent \textbf{Results} From Figure \ref{fig:decsur}(a), we see that the model gets confused whenever the animal's face is not visible, or if it is in a position facing away from the camera. This is reasonable since we are only measuring the most confusing samples between the \textit{Tabby Cat} and \textit{Great Dane} classes which share a lot of semantic similarity. Similarly, in the MNIST dataset, the examples shown depict atypical ways in which the digits `0' and `6' can be written. These results suggest that MARGIN is effective in identifying these examples, with a combination of the appropriate neighborhoods (in the latent space of the model) and labels. 
\vspace{10pt}

\subsection{Case Study V - Characterizing Statistics of Adversarial Examples}
\label{sec:adversarial}
\begin{figure*}[!h]
  \centering
    \includegraphics[width=.95\linewidth,clip=True,trim=0 0 0 0]{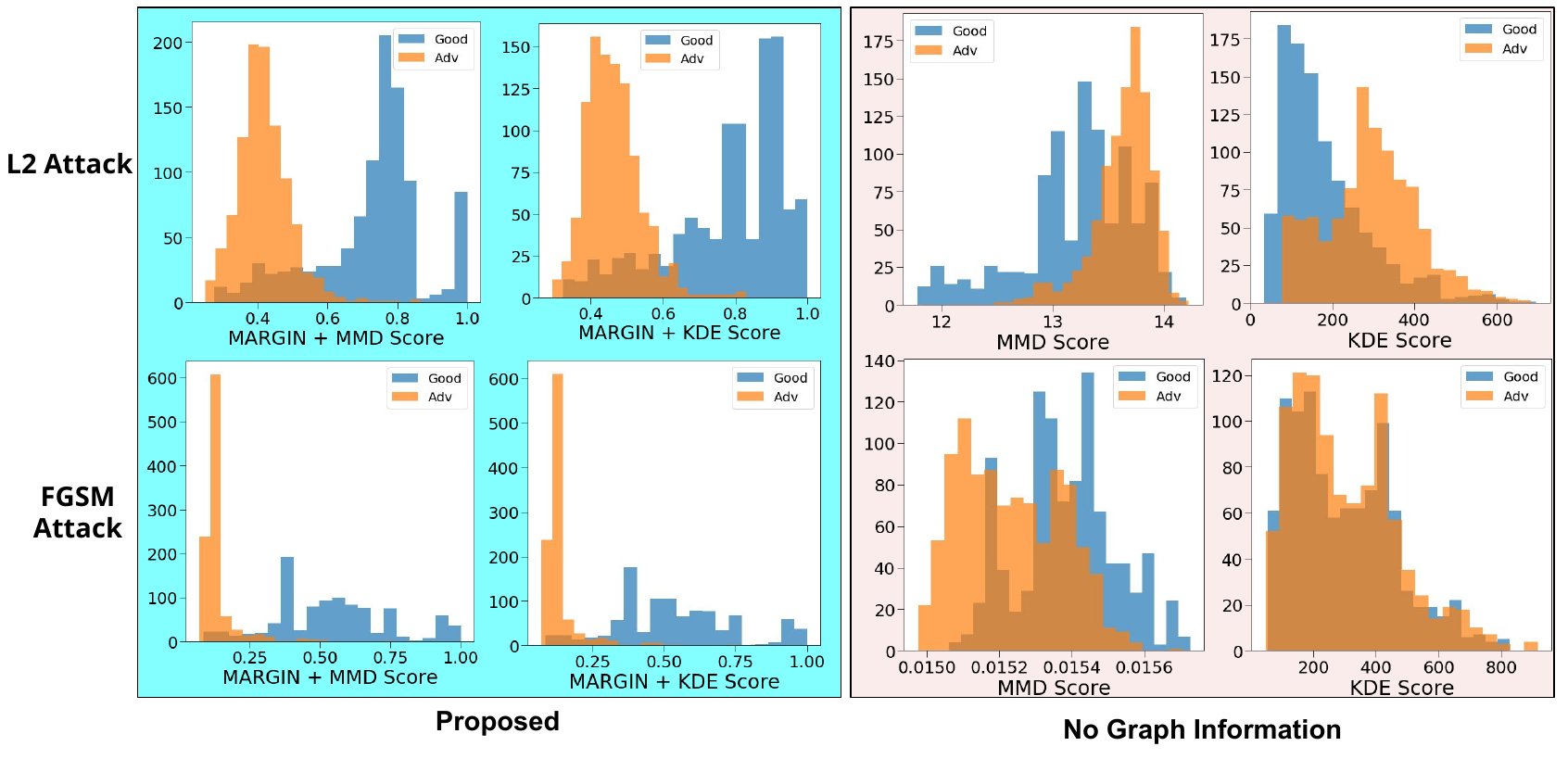}
\caption{We compare histograms of scores obtained from adversarial samples with and without incorporating graph structure. We see that including the structure results in a much better separation between adversarial and harmless examples. In addition, regions of overlap can easily be explained.}
\label{fig:adv}
\end{figure*}

In this application, we examine the problem of quantifying the statistical properties of adversarial examples using MARGIN. Adversarial samples \citep{biggio2013evasion,szegedy2013intriguing} refer to examples that have been specially crafted, such that a particular trained model is `tricked' into mis-classifying them. This is done typically by perturbing a sample, sometimes in ways imperceptible to humans, while maximizing mis-classification rates. In order to better understand the behaviour of such adversarial examples, there have been studies in the past to show that adversarial examples are statistically different from normal test examples. For example, an MMD score between distributions is proposed in \citep{grosse2017statistical}, and a kernel density estimator (KDE) in \citep{feinman2017detecting}. However, these measures are global, and provide little insight into individual samples. We propose to use \name to develop these statistical measures at a sample level, and study how individual adversarial samples differ from regular samples.

\noindent \textbf{Formulation:} As in other case studies, MARGIN constructs a graph, where each node corresponds to an example that is either adversarial or harmless, and the edges are constructed using neighbors in the latent space of the model, against which the adversarial examples have been designed. We consider two kinds of functions in this experiment: i) \textbf{MMD Global:} Similar to \ref{sec:critics}, we use the MMD score between the whole set, and the set without a particular sample and its neighbors. This provides a way to capture statistically rarer samples in the dataset; ii) \textbf{KDE:} We also use the KDE of each sample, as proposed in \citep{feinman2017detecting}, where we measure the discrepancy of each sample against the training samples from its predicted class. While these measures on their own may not be very illustrative, they are useful functions to determine influences within MARGIN.

\noindent \textbf{Experiment Setup and Results:} We perform experiments on $2000$ randomly sampled test images from the MNIST dataset \citep{lecun1998mnist}, of which we adversarially perturb $1000$ images. We measure MARGIN scores using both MMD Global, and KDE, against two popular attacks -- the Fast Gradient Sign Method (FGSM) attack \citep{goodfellow2014explaining}, and the L2-attack \citep{carlini2017towards}. We use the same setup as in \citep{carlini2017adversarial}, including the network architecture for MNIST. The resulting \name score determined using algorithm \ref{alg:select} is more discriminative, as seen in Figure \ref{fig:adv}. As noted in \citep{carlini2017adversarial}, the MMD and KDE measures were not very effective against stronger attacks such as the L2-attack. This is reflected to a much lower degree even in our approach, where there is a small overlap in the distributions. We also find that the overlapping regions correspond to samples from the training set that are extremely rare to begin with (like criticisms from section \ref{sec:critics}).

\subsection{Case Study VI - Active Learning on Graphs}
To demonstrate the applicability of MARGIN to work with graph structured data, we study the problem of active learning on graphs, or in other words, generating highly influential samples for a label propagation task. Label propagation is a semi-supervised learning problem, where the task is to propagate labels from a small set of nodes to all the other nodes of the graph. In order to evaluate the samples chosen using our method, we study the test accuracies for varying sizes of the training set. In order to perform the semi-supervised learning, we use the Graph Convolutional Network (GCN) implementation by \citep{Kipf2016GCNN}, with 3 graph convolutional layers comprising 16 graph filters each, and a learning rate of 0.01. The rest of the hyper-parameters are those recommended in the GCN implementation\footnote{\url{https://github.com/tkipf/gcn}}.

\noindent \textbf{Formulation} Since the attributes are independently defined on each node, they do not contain information about the neighborhoods in the graph and therefore do not directly provide us a notion of influence. Instead, we first embed the attributes using a graph convolutional autoencoder \citep{kipf2016variational}, and compute the explanation function $\mathbf{f}$ as the as the norm of each latent feature at each node. Next, using MARGIN we compute the influences of the training samples alone, and sort them in decreasing order. 
\noindent \textbf{Datasets and Baselines:} We consider two popularly used citation network datasets -- Cora and Citeseer \citep{sen2008collective}. The Cora dataset consists of 2,708 nodes and 5,429 edges, while the Citeseer dataset consists of 3,327 nodes and 4,732 edges. The attributes at each node are comprised of a sparse bag-of-words feature vector with 3,703 dimensions for Citeseer, and 1,433 dimensions for Cora. 

\begin{figure}[!htb]
	\centering
	\includegraphics[width=.95\linewidth,clip=True]{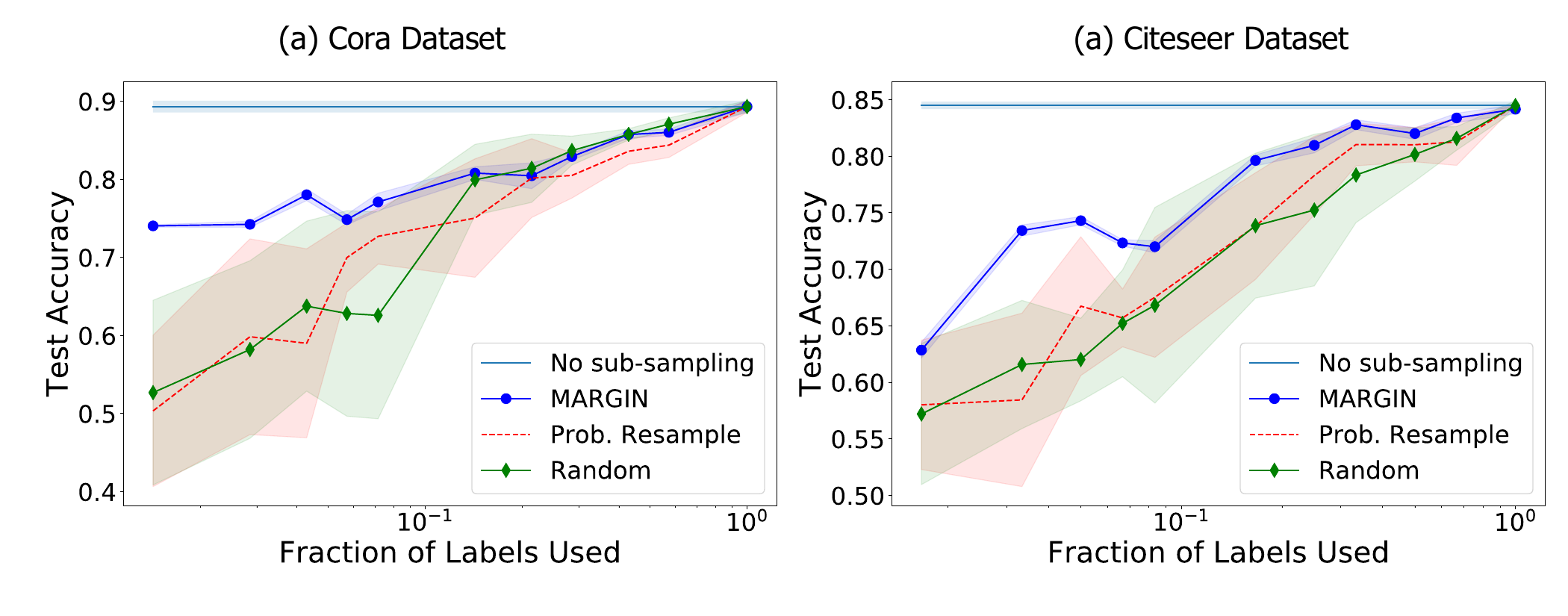}
	\caption{MARGIN based sampling for graph signals shows improvement in label propagation performance, even for very small sets of samples. Here we perform label propagation using a very small set of training nodes to the rest of the graph in each dataset shown.}
	\label{fig:gcn}
\end{figure}

\noindent We compare with two baselines: {\it (i) Probabilistic resampling on graphs:} The resampling strategy was proposed in \citep{chen2017fast} as a way to efficiently resample dense point clouds. In this approach, the magnitude of the features at each node after a high pass filtering is directly used as a probability of influence at that node, $p(n)$. This is followed by a resampling of the nodes on the graph according to $p(n)$. While it is an effective strategy to resample dense point clouds, it tends to be less reliable for the label propagation experiment, as shown in Figure \ref{fig:gcn}.  {\it (ii) Random sampling:} We also randomly sample from each class on the graph. In all experiments, we repeat the sampling 5 times, and retrain the GCN each time with the set of nodes to report the mean and standard deviation.

\vspace{0.5em}

\noindent {\bf Results:} In all cases, the accuracy of label propagation is measured on a test set of size 1,000 samples, by training on only 10-100s of samples. Figure \ref{fig:gcn} shows the accuracy of label propagation for varying number of training set sizes. It is clear that our proposed sampling achieves state-of-the art performance on the graph. The performance is around $15–20\%$ points higher in accuracy compared to the baseline techniques, especially in small training set regimes. We repeat all the techniques 5 times with different random seeds and show the mean and standard deviation. In the experimental setup, we sample from as few as 1 node per class, to using all the samples in each class for label propagation onto the nodes not used in training. Here, the total training nodes are $140$ for Cora and $120$ for Citeseer. The non-monotonic behaviour is partly owing to the fact that at each step we perform a new experiment by sampling a new set of $N_{train}$ nodes (instead of augmenting existing nodes) for all methods in consideration, and from figure \ref{fig:gcn} it is clear that MARGIN is able to identify the most important nodes for propagating the label information when compared to baselines. Further, the probabilistic resampling method also acts as an ablation on MARGIN where we use the same function but without the high frequency filtering, illustrating that MARGIN is critical in obtaining influence, instead of simply using the function as a probability measure, as done here. Finally, random sampling itself is a competitive baseline as the number of samples under consideration is very small, yet has a very high variation in the prediction performance.

\vspace{5pt}

\subsection{Scalability of MARGIN}
Large scale datasets and multi-billion parameter models are increasingly becoming the norm in deep learning, bringing the issue of scalability in interpretable ML to the forefront.  So far, this has received relatively lesser attention in part because it is unclear if interpreting at scale is the ultimate goal since most explanations tend to be local -- i.e. with respect to a single sample or image, or a small neighborhood of samples. Regardless, some approaches tend to be inherently more scalable than others during inference, for example CXPlain \citep{schwab2019cxplain}, ProFile \citep{thiagarajan2020accurate}, DeepSHAP \citep{shrikumar2017learning}, while others like LIME are computationally more expensive \citep{ribeiro2016should}. While these methods can explain efficiently, they are limited by the scope of interpretability applications that can be addressed, while MARGIN is more broadly applied while being more expensive to compute. Since MARGIN operates by filtering a graph signal, the main bottlenecks for scalability are in the graph construction step, and need to store massive graphs in memory for filtering. Techniques developed in the last few years have shown an impressive ability to scale the problem of graph construction to millions of nodes \citep{liu2019scalable}, while learning methods on graphs have also been scaled similarly \citep{ying2018graph}. These open avenues of future work that enable scaling MARGIN to massive datasets so it may also be applied at web-scale. 

\section{Discussion and Conclusion}
We proposed a generic framework called MARGIN that is able to provide explanations to popular interpretability tasks in machine learning. These range from identifying prototypical samples in a dataset that might be most helpful for training, to explaining salient regions in an image for classification. In this regard, MARGIN exploits ideas rooted in graph signal processing to identify the most influential nodes in a graph, which are nodes that maximally affect the graph function. While the framework is extremely simple, it is highly general in that it allows a practitioner to include rich semantic information easily in three crucial ways -- defining the domain (intra-sample vs inter-sample), edges (pre-defined/native/model latent space), and finally a function defined at each node. The case studies in this paper have focused on use cases that are most commonly encountered in the real world.

\newpage
\section*{Python Implementation of MARGIN}
\label{code:margin}
The graph analysis based influence estimation in MARGIN is extremely simple, in that it can be implemented using a few lines of python code.

\begin{python}
import numpy as np
import networkx as nx
import scipy.sparse as sp
'''
Inputs:
adj : adjacency matrix
f : function defined at each node
p : number of hops from each node
    for filtering
Output:
I : Influence score per node
'''
def MARGIN(adj,f,p=1):
    G = nx.Graph(adj) #graph object
    N = adj.shape[0] # number of nodes
    degree =  dict(G.degree()) 

    deg = [1./d[1] for d in degree.items()]
    Dinv = sp.csr_matrix(np.zeros((N,N)))
    idx0,idx1 = np.diag_indices(N)
    Dinv[idx0,idx1] = deg #degree matrix
    
    A_norm = np.sqrt(Dinv)*adj*np.sqrt(Dinv)
    P = A_norm**p #multiple-hops

    M = np.sum(P>0,axis=1,dtype=np.float)

    f_filter = f-(np.matmul(P.todense(),f))/M
    I = np.abs(f_filter)
    I = I/np.max(I)
    
    return I.A

  \end{python}

\section*{Acknowledgement}
This work was performed under the auspices of the U.S. Dept. of Energy by Lawrence Livermore National Laboratory under Contract DE-AC52-07NA27344.

\section*{Disclaimer}
\noindent This document was prepared as an account of work sponsored by an agency of the United States government. Neither the United States government nor Lawrence Livermore National Security, LLC, nor any of their employees makes any warranty, expressed or implied, or assumes any legal liability or responsibility for the accuracy, completeness, or usefulness of any information, apparatus, product, or process disclosed, or represents that its use would not infringe privately owned rights. Reference herein to any specific commercial product, process, or service by trade name, trademark, manufacturer, or otherwise does not necessarily constitute or imply its endorsement, recommendation, or favoring by the United States government or Lawrence Livermore National Security, LLC. The views and opinions of authors expressed herein do not necessarily state or reflect those of the United States government or Lawrence Livermore National Security, LLC, and shall not be used for advertising or product endorsement purposes.

\bibliography{refs}
\bibliographystyle{ieee}

\end{document}